\documentclass[11pt, a4paper]{article}
\pdfoutput=1 

\usepackage[a4paper, top=2.5cm, bottom=2.5cm, left=2cm, right=2cm]{geometry}
\usepackage[T1]{fontenc}
\usepackage[english]{babel}
\usepackage{lmodern}
\usepackage{amsmath}
\usepackage{amssymb}
\usepackage{booktabs}
\usepackage{array}
\usepackage{caption}
\usepackage{enumitem}
\usepackage{graphicx}
\usepackage[colorlinks=true, linkcolor=blue, citecolor=blue, urlcolor=blue]{hyperref}

\usepackage{listings}
\usepackage{xcolor}

\definecolor{codegreen}{rgb}{0,0.6,0}
\definecolor{codegray}{rgb}{0.5,0.5,0.5}
\definecolor{codepurple}{rgb}{0.58,0,0.82}
\definecolor{backcolour}{rgb}{0.95,0.95,0.92}

\lstdefinestyle{gnwmstyle}{
    backgroundcolor=\color{backcolour},   
    commentstyle=\color{codegreen},
    keywordstyle=\color{blue},
    numberstyle=\tiny\color{codegray},
    stringstyle=\color{codepurple},
    basicstyle=\ttfamily\footnotesize,
    breakatwhitespace=false,         
    breaklines=true,                 
    captionpos=t,                    
    keepspaces=true,                 
    numbers=none,                    
    numbersep=5pt,                  
    showspaces=false,                
    showstringspaces=false,
    showtabs=false,                  
    tabsize=4,
    frame=lines
}
\setlist[itemize]{label=-}
\title{The Global Neural World Model: Spatially Grounded Discrete Topologies for Action-Conditioned Planning}
\author{Noureddine Kermiche \\ \small{Western Digital Corporation, Irvine, CA, USA}}
\date{}

\begin{document}

\maketitle

\begin{abstract}
We present the Global Neural World Model (GNWM), a self-stabilizing framework that achieves topological quantization through balanced continuous entropy constraints. Operating as a continuous, action-conditioned Joint-Embedding Predictive Architecture (JEPA), the GNWM maps environments onto a discrete 2D grid, enforcing translational equivariance without pixel-level reconstruction. Our results show this architecture prevents manifold drift during autoregressive rollouts by using grid ``snapping'' as a native error-correction mechanism. Furthermore, by training via maximum entropy exploration (random walks), the model learns generalized transition dynamics rather than memorizing specific expert trajectories. We validate the GNWM across passive observation, active agent control, and abstract sequence regimes, demonstrating its capacity to act not just as a spatial physics simulator, but as a causal discovery model capable of organizing continuous, predictable concepts into structured topological maps.
\end{abstract}

\vspace{0.5cm}
\vspace{0.5cm}

\section{Introduction}
The pursuit of artificial agents capable of long-term planning relies heavily on the development of accurate, persistent world models [5, Ha \& Schmidhuber, 2018]. However, current state-of-the-art predictive architectures predominantly rely on continuous latent representations. While continuous spaces are efficient for gradient-based learning, they introduce two fundamental pathologies in temporal models: interpretability opacity and iterative drift.

In autoregressive rollout scenarios, continuous models inevitably succumb to the ``mean-prediction'' problem. Small floating-point inaccuracies at step $t+1$ compound exponentially, causing the predicted state to diffuse into high-entropy regions of the latent space. This manifests as visual blur and a rapid loss of physical groundedness.

To bridge the gap between continuous perception and discrete logic, we propose the Global Neural World Model (GNWM). The GNWM treats the latent space as a discrete topological grid governed by balanced entropy constraints. Inspired by the topological ordering of classic Self-Organizing Maps (SOMs)---which historically resisted deep learning integration due to non-differentiable operations---the GNWM introduces a fully differentiable method for unrolling an environment's causal manifold.

Our primary contributions are:
\begin{enumerate}
    \item \textbf{Self-Stabilizing Entropy Constraints:} A balanced objective function that organically prevents representation collapse on a probability simplex, removing the need for asymmetric gradient routing, explicit variance penalties, or target networks.
    \item \textbf{Fully Differentiable Topological Smearing:} We solve the historical incompatibility between SOMs and deep learning by replacing Kohonen's non-differentiable Winner-Take-All spike and heuristic radial weight updates with a fixed, fully differentiable spatial convolution. This decouples topology from activation, allowing exact gradient flow.
    \item \textbf{Emergent Topological Disentanglement:} Demonstration that 2D SOM convolutions inherently force continuous features to geometrically disentangle across discrete Cartesian axes.
    \item \textbf{Drift-Free Imagination:} Empirical proof that discrete topological quantization allows for autoregressive planning over extended horizons without variance decay.
    \item \textbf{Compositional Factorization \& Cognitive Mapping:} Demonstration of architectures that organically separate multi-entity environments into independent grids, and successfully cluster abstract semantic concepts into structured topological dictionaries.
\end{enumerate}

\section{Related Work}
The GNWM builds upon and diverges from four primary paradigms in representation learning:

\subsection{Continuous Predictive Models and Anti-Collapse Heuristics}
Non-contrastive predictive architectures like JEPAs [1, Bardes et al., 2024] and BYOL [4, Grill et al., 2020] aim to predict future states in continuous space. However, without explicit repulsive forces, these models inevitably succumb to representational collapse. To prevent this, they resort to a fragile stack of optimization heuristics: architectural asymmetry (Exponential Moving Average target networks with stop-gradients), heavy weight decay, and strict batch or variance normalization. 

Conversely, Contrastive Learning methods (e.g., SimCLR [3, Chen et al., 2020]) explicitly prevent collapse by pushing negative samples apart, but they scale quadratically with batch size ($\mathcal{O}(D \cdot N^2)$). Covariance whitening methods (e.g., VICReg [2, Bardes et al., 2021], Barlow Twins [8, Zbontar et al., 2021]) mitigate batch dependencies by decorrelating dimensions, but they scale quadratically with the embedding dimension ($\mathcal{O}(D^2 \cdot N)$). Both approaches introduce severe computational bottlenecks. The GNWM circumvents this entire optimization stack. By formulating representation learning as a balanced thermodynamic equilibrium, the GNWM is mathematically self-stabilizing in strictly linear time.

\subsection{Discrete Latent Variables (VQ-VAEs)}
Vector Quantized models project inputs into a discrete codebook. However, they struggle with the non-differentiability of the hard \texttt{argmax} step, relying on the Straight-Through Estimator. This often leads to codebook collapse. The GNWM replaces the STE with continuous differentiable probabilities during training, utilizing hard quantization only during inference.

\subsection{Self-Organizing Maps (SOMs)}
Kohonen's original SOMs [7, Kohonen, 1982] pioneered mapping high-dimensional data onto ordered grids but completely resisted integration into the modern deep learning revolution. This incompatibility stems from a fundamental flaw in the classic algorithm: it operates via a non-differentiable two-step process. First, it relies on a hard \texttt{argmin} to locate a Best Matching Unit (BMU), which instantaneously severs the computational graph. Second, it utilizes a Radial Basis Function (RBF) as a heuristic weight-update modifier (effectively a dynamic learning rate) rather than a mathematical operation within the forward pass, making standard backpropagation impossible. 

The GNWM revitalizes the SOM paradigm by fundamentally altering its mechanics. We replace the hard WTA spike with continuous thermodynamic entropy losses, and critically, we replace the heuristic radial weight update with a fixed spatial convolution applied directly to a continuous logit grid. This guarantees perfect end-to-end differentiability.

\subsection{Self-Organizing Representation Learning}
Recent advances [6, Kermiche, 2021] demonstrated mapping datasets onto 2D grids using continuous embeddings and Gaussian kernels. The GNWM extends this static clustering into the temporal domain to create autoregressive world models.

\section{Mathematical Framework}

\begin{figure}[htbp]
    \centering
    \includegraphics[width=\textwidth]{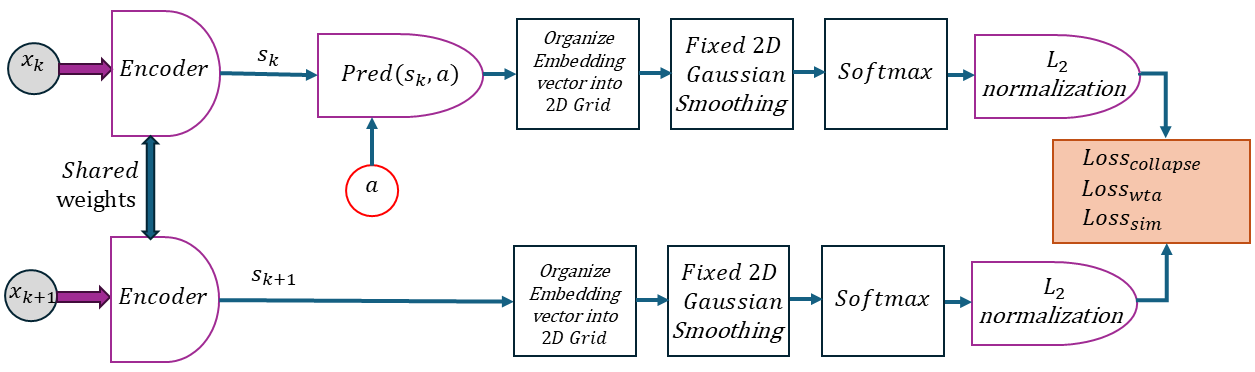}
    \caption{The GNWM Architecture. Raw inputs are processed by the Retinotopic Encoder into a spatial grid. A fixed radial filter applies Topological Smearing. The Action-Conditioned Spatial Predictor then estimates the future latent state, enforcing a strict JEPA paradigm.}
    \label{fig:architecture_gnwm}
\end{figure}

\subsection{The Topological Probability Simplex}
Let the latent space be defined as a 2D grid of size $H \times W$, containing $D = H \times W$ discrete neural nodes. For a given input state $x_t$, the encoder produces a continuous latent vector $h_t \in \mathbb{R}^D$.

To enforce topological continuity without breaking the gradient, we apply a 2D SOM convolution to the latent activations via a fixed Gaussian kernel $G_\sigma$ centered at each node:
\begin{equation}
\tilde{h}_t = h_t \ast G_\sigma
\end{equation}

By treating the radial filter as a structural, fixed spatial convolution rather than a dynamic activation function, gradients flow perfectly back through the neighborhood spread. Furthermore, this operation introduces ``Topological Smearing.'' Instead of a hard spike that causes activation mass to abruptly ``teleport'' between discrete nodes across frames, the probability mass smoothly transitions and smears across the grid. This allows the predictive alignment loss ($\mathcal{L}_{sim}$) to natively optimize for continuous physical motion over a discrete topology. 

This topologically smoothed vector is then projected onto the probability simplex and $L_2$-normalized to create a spherical probability distribution:
\begin{equation}
p_t = \frac{\text{Softmax}(\tilde{h}_t)}{\|\text{Softmax}(\tilde{h}_t)\|_2}
\end{equation}

As seen in Figure 1, both the predicted state $p_{t+1}$ and target state $z_{t+1}$ undergo this exact topological transformation. Crucially, the target state is defined strictly as $z_{t+1} = \text{Encoder}(x_{t+1})$. While the forward pathways are inherently asymmetric---since the Spatial Predictor operates on a latent state modified by an action vector $a_t$, whereas the target branch solely processes raw future frames---the Encoder utilizes exact shared weights across all time steps. This shared-weight encoding mathematically proves that the GNWM's thermodynamic equilibrium prevents representation collapse organically, entirely eliminating the need for the asymmetric Exponential Moving Average (EMA) target networks and stop-gradients required by standard JEPA architectures.

\subsection{Thermodynamic Equilibrium: Expansion vs. Contraction}

\begin{figure}[htbp]
    \centering
    \includegraphics[width=\textwidth]{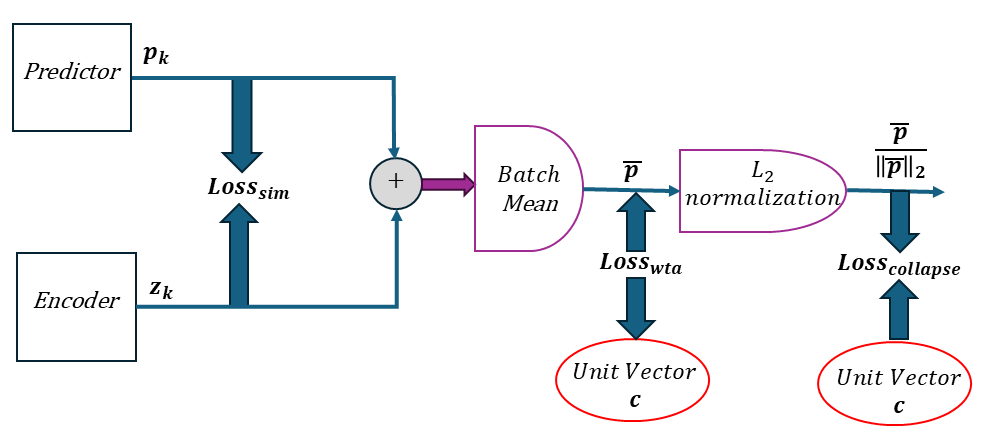}
    \caption{The Thermodynamic Gradient Flow. The batch mean is evaluated against the uniform constant ($c$) strictly on the target state, naturally balancing expansion and contraction forces }
    \label{fig:thermo}
\end{figure}

Let $\bar{p} = \frac{1}{B} \sum_{i=1}^B \frac{1}{2} (p^{(i)} + z^{(i)})$ represent the unnormalized global batch mean. We introduce two competing loss functions evaluated against the uniform constant vector $c = \frac{\mathbf{1}}{\sqrt{D}}$ (see Figure 2).

\textbf{The Expansion Force (Collapse Loss):}
\begin{equation}
\mathcal{L}_{collapse} = 1 - \left\langle \frac{\bar{p}}{\|\bar{p}\|_2}, c \right\rangle
\end{equation}
Minimizing this forces the network to utilize the entire grid uniformly across the batch.

\textbf{The Contraction Force (Winner-Take-All Loss):}
\begin{equation}
\mathcal{L}_{WTA} = \langle \bar{p}, c \rangle
\end{equation}
Minimizing this acts as an entropic contraction force, penalizing blur and forcing highly peaked states.

\textbf{Computational Efficiency:}
Crucially, because the GNWM's thermodynamic mechanism evaluates the batch mean against a fixed uniform constant vector ($c$) rather than computing pairwise sample distances or covariance matrices, the entire collapse-prevention system scales perfectly linearly with both batch size and embedding dimension ($\mathcal{O}(D \cdot N)$). This fundamentally bypasses the $\mathcal{O}(D \cdot N^2)$ and $\mathcal{O}(D^2 \cdot N)$ computational bottlenecks of traditional contrastive and whitening methods, while completely eliminating the need for BYOL-style weight decay or batch normalization heuristics.

\subsubsection{Proof of the Global Minimum for Arbitrary Dimension $D$}

Let the combined thermodynamic equilibrium be defined strictly as the sum of the expansion and contraction forces: $\mathcal{L}_{thermo} = \mathcal{L}_{collapse} + \mathcal{L}_{WTA}$. Building upon symmetry arguments for 2-neuron systems [6, Kermiche, 2021], we can mathematically prove that for any arbitrary dimension $D$, the global minimum of this equilibrium is uniquely achieved when representations are perfectly quantized and uniformly distributed.

For any $L_2$-normalized positive vector, the $L_1$ norm is bounded by $1 \le \|p\|_1 \le \sqrt{D}$. The lower bound of $1$ is strictly achieved if and only if $p^{(i)}$ is a one-hot state. Therefore, $\mathcal{L}_{WTA} \ge \frac{1}{\sqrt{D}}$. 
Conversely, by the Cauchy-Schwarz inequality, $\mathcal{L}_{collapse}$ achieves its maximum (loss = $0$) if and only if the global batch average is perfectly uniform.
Therefore, the absolute global minimum ($\mathcal{L}_{thermo} \ge \frac{1}{\sqrt{D}}$) is attainable if and only if the network outputs perfect one-hot vectors and distributes them equally across all $D$ grid nodes.

\subsection{Predictive Alignment (Similarity Loss)}
The network learns transition dynamics by aligning the autoregressive prediction $p_{t+1}$ with the target $z_{t+1}$:
\begin{equation}
\mathcal{L}_{sim} = 1 - \frac{1}{B} \sum_{i=1}^B \langle p_{t+1}^{(i)}, z_{t+1}^{(i)} \rangle
\end{equation}

\subsection{The Unified Objective Function}
The final loss function strictly achieves self-organization through internal thermodynamic unrolling and predictive alignment:
\begin{equation}
\mathcal{L}_{total} = \alpha (\mathcal{L}_{collapse} + \mathcal{L}_{WTA}) + \gamma \mathcal{L}_{sim}
\end{equation}
where $\alpha$ serves as a dynamically scheduled heating parameter during early epochs.The entire objective reduces to a few lines of highly optimized, linear tensor operations as shown in Listing 1.

\begin{lstlisting}[language=Python, caption=PyTorch implementation of Geometric Self-Organization, label=lst:geometric_self_org]
# p, z: SOM-convolved raw logits for predictor and target (Batch, D)
const = F.normalize(torch.ones(1, D, device=device), p=2, dim=1)

# 1. Simplex & Hypersphere Projection
p_l2 = F.normalize(F.softmax(p, dim=1), p=2, dim=1)
z_l2 = F.normalize(F.softmax(z, dim=1), p=2, dim=1)

# 2. Global Batch-Mean State
mean_vec = torch.mean((p_l2 + z_l2) / 2.0, dim=0, keepdim=True)
mean_l2 = F.normalize(mean_vec, p=2, dim=1)

# 3. Thermodynamic Loss
L_collapse = 1.0 - torch.sum(mean_l2 * const)
L_WTA = torch.sum(mean_vec * const)
L_sim = 1.0 - torch.mean(torch.sum(p_l2 * z_l2, dim=1))
alpha = 1.0
gamma = 0.5

loss = alpha * (L_collapse + L_WTA) + gamma * L_sim
\end{lstlisting}

\subsection{The Inseparability of the Thermodynamic System}
These geometric components are strictly inseparable. Without $\mathcal{L}_{collapse}$, representations trivially collapse. Without $\mathcal{L}_{WTA}$, predictions revert to conditional blur. Without $\mathcal{L}_{sim}$, the network ignores temporal sequences. Without Topological Smoothing ($G_\sigma$), the network activates random disjoint neurons, shattering local geometry. PyTorch implementation of Topological Smoothing is shown in Listing 2.

\begin{lstlisting}[language=Python, caption=Neighborhood Recruitment in PyTorch, label=lst:neighborhood_recruitment]
# 1. Define fixed topological constraint
sigma = 1.5 
k = int(6 * sigma) | 1  # Ensure an odd kernel size

# 2. Generate 2D Gaussian kernel
grid = torch.arange(-k//2 + 1, k//2 + 1, dtype=torch.float32)
y_grid, x_grid = torch.meshgrid(grid, grid, indexing='ij')
gaussian = torch.exp(-(x_grid**2 + y_grid**2) / (2*sigma**2))
w = (gaussian / torch.sum(gaussian)).view(1, 1, k, k)

# 3.Apply radial smoothing to latent grid h:(Batch,1,H,W)
h_smoothed = F.conv2d(h, w.to(h.device), padding='same')
h_flat = h_smoothed.view(h.size(0), -1)
\end{lstlisting}

\section{Architecture}

\subsection{Retinotopic Visual Encoder}
To preserve 2D geometry, the GNWM utilizes a Fully Convolutional Encoder mapping raw RGB frames directly to a spatial logit grid. By omitting dense linear bottlenecks, the physical top-left of the visual field is mathematically locked to the top-left of the latent grid, allowing the network to factorize complex scenes into separate, orthogonal grids. For abstract, non-spatial sequence tasks (e.g., Paradigm D), this convolutional stack is seamlessly replaced by a standard Multi-Layer Perceptron (MLP) to project flat continuous embeddings onto the 2D grid, preserving the thermodynamic constraints without requiring 2D visual arrays.

\subsection{Spatial Transition Predictor (Action-Conditioned JEPA)}
The transition model operates directly on the probability simplex, functioning as an action-conditioned Joint-Embedding Predictive Architecture (JEPA). Rather than relying on pixel-level decoders, the GNWM predicts the latent representation of the future state. The discrete action vector is broadcast across the latent state and processed via spatial convolutions. This architectural choice enforces translational equivariance: the causal logic of ``moving right'' applies the same spatial shift to the probability mass regardless of the agent's absolute location on the grid. 

Crucially, to learn a robust world model, the Predictor must not overfit to the narrow trajectories of a specialized expert policy. By training the architecture on data generated via maximum entropy exploration (random actions), we force the GNWM to map a broader state-action distribution. This ensures the model learns the underlying transition dynamics---including edge cases and boundary collisions---rather than merely memorizing optimal paths, which helps mitigate hallucination during subsequent planning rollouts.

\section{Experimental Setup}

 \textbf{GitHub} repository containing simulation scripts can be found \href{https://github.com/norikermiche-123/GNWM}{here.}

\textbf{5.1 Paradigm A: Passive Observation} \\
A 1,200-frame video of a single ball moving with fixed momentum. The GNWM must passively organize the states into the discrete $15 \times 15$ grid without action labels.

\textbf{5.2 Paradigm B: Active Agent Control} \\
A single ball's trajectory driven entirely by discrete, randomized agent actions (Up, Down, Left, Right). The spatial predictor receives the current state and action to predict the subsequent frame.

\textbf{5.3 Paradigm C: Compositional Multi-Object Factorization} \\
An environment containing two independently bouncing balls (Red and Blue). The Retinotopic Encoder outputs a dual-channel manifold, computing thermodynamic constraints independently for each channel to evaluate dimensionality reduction.

\textbf{5.4 Paradigm D: Abstract Semantic Topology} \\
Having established the architecture's capacity for physical kinematics and multi-object factorization, we hypothesized that these spatial thermodynamics could act as a generalized causal mapping engine for any predictable sequence. To demonstrate this, we constructed a synthetic language modeling task with 40 unique 32D word embeddings governed by a rigid grammatical rule (Noun $\rightarrow$ Verb $\rightarrow$ Adjective $\rightarrow$ Object). 

\section{Experimental Results}

\subsection{Semantic State Space and Grid Utilization}

\begin{figure}[htbp]
    \centering
     \includegraphics[width=0.5\textwidth]{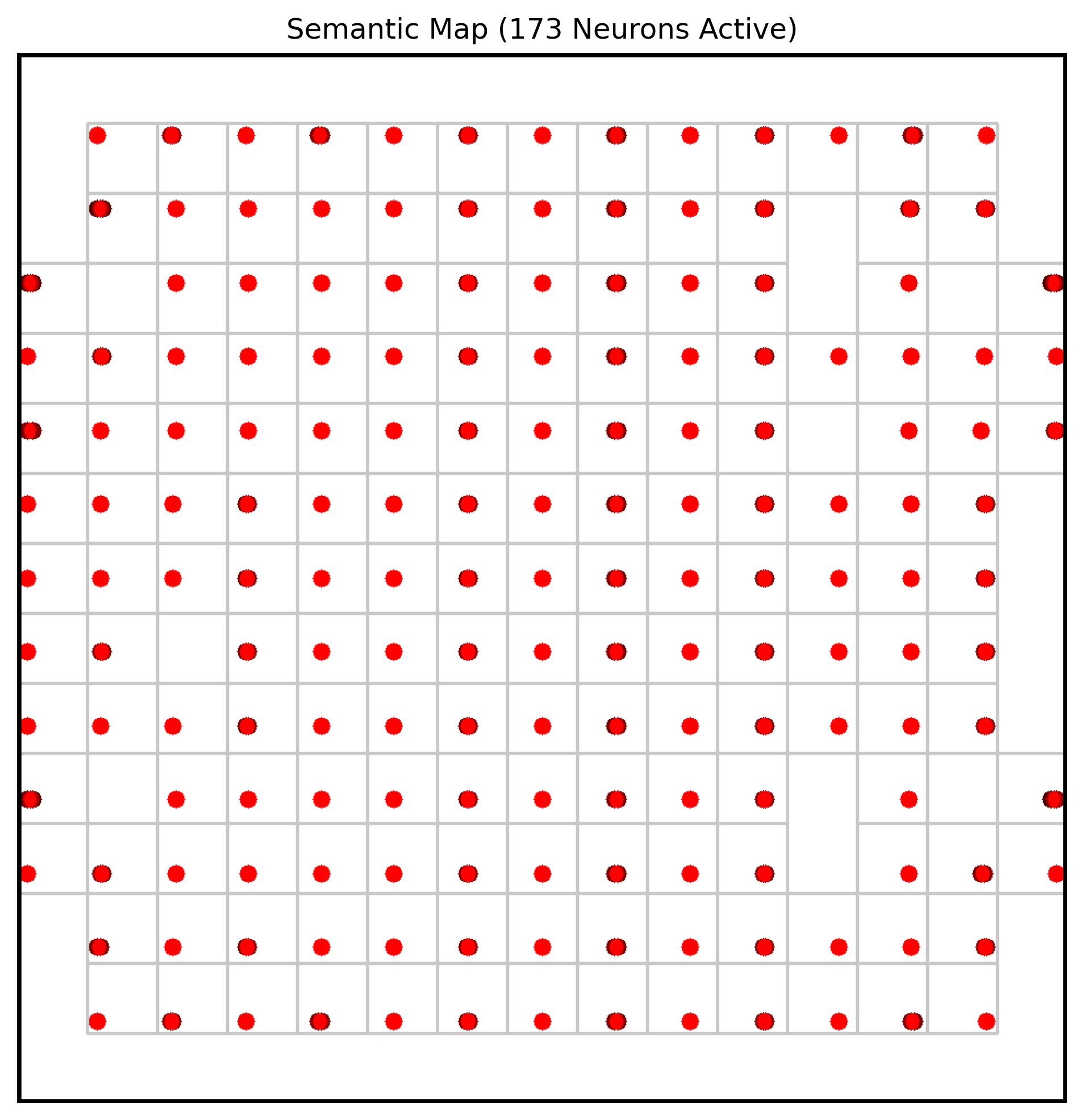}
    \caption{Visual centroids mapping the continuous bouncing ball environment into an interpretable, discrete stop-motion dictionary without heuristic labels.}
    \label{fig:dictionary}
\end{figure}

In the passive observation task, the GNWM successfully mapped the environment without codebook collapse. Relieved of artificial variance constraints, the active neurons organically expanded to form a contiguous block, utilizing 173 of the 225 available neurons, creating a spatially ordered ``stop-motion'' dictionary as seen in Figure 3.

To visualize the learned representations, we project the continuous video frames onto the discrete topological grid by identifying the Winner-Take-All (WTA) Best Matching Unit for each frame. By computing the mean average of all frames associated with a single discrete neuron, we can render its underlying \textit{prototype state}. Importantly, the neurons do not simply memorize static, isolated frames. Instead, they capture generalized conceptual manifolds---manifesting as continuous "ghosted" trails that compress dynamic spatiotemporal concepts (e.g., "object moving rightward in the upper quadrant") into a single, drift-free discrete node.

\subsection{Drift-Free Autoregressive Rollouts}

\begin{figure}[htbp]
    \centering
    \includegraphics[width=0.7\textwidth]{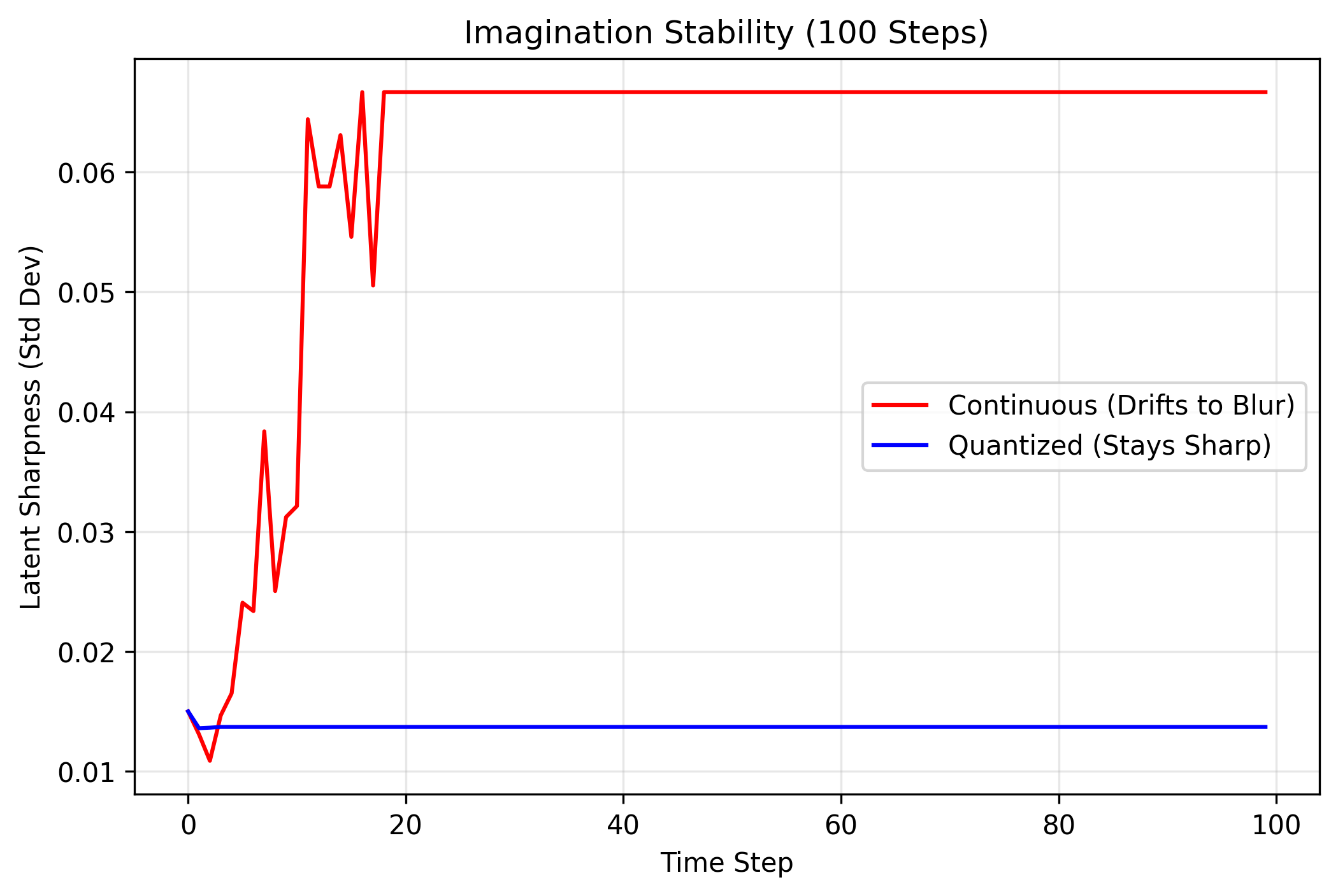}
    \caption{Latent variance over a 100-step horizon. Grid snapping prevents the mean-prediction blur common in continuous architectures.}
    \label{fig:variance}
\end{figure}

Evaluated over a 100-step autoregressive horizon, continuous baselines decayed to a high-entropy mean (standard dev $0.066$). By utilizing ``grid snapping''---an inference-time operation where the continuous predicted probability mass $p_{t+1}$ is replaced by a strict one-hot vector located at its \texttt{argmax} coordinate before the next recurrent step---the GNWM maintained constant physical sharpness ($0.016$) throughout the entire rollout, acting as a native Error-Correction Code as seen in Figure 4.

\subsection{Action-Conditioned Imagination Trees}
In the active control paradigm, the Spatial Predictor successfully generated deterministic, orthogonal branches across the latent grid for all 4 available actions. The network autonomously allocated exactly 41 active neurons, as seen in Figure 5, perfectly matching the empirical visitation distribution of the active random walk without hallucinating dead nodes.

\begin{figure}[htbp]
    \centering
    \includegraphics[width=0.5\textwidth]{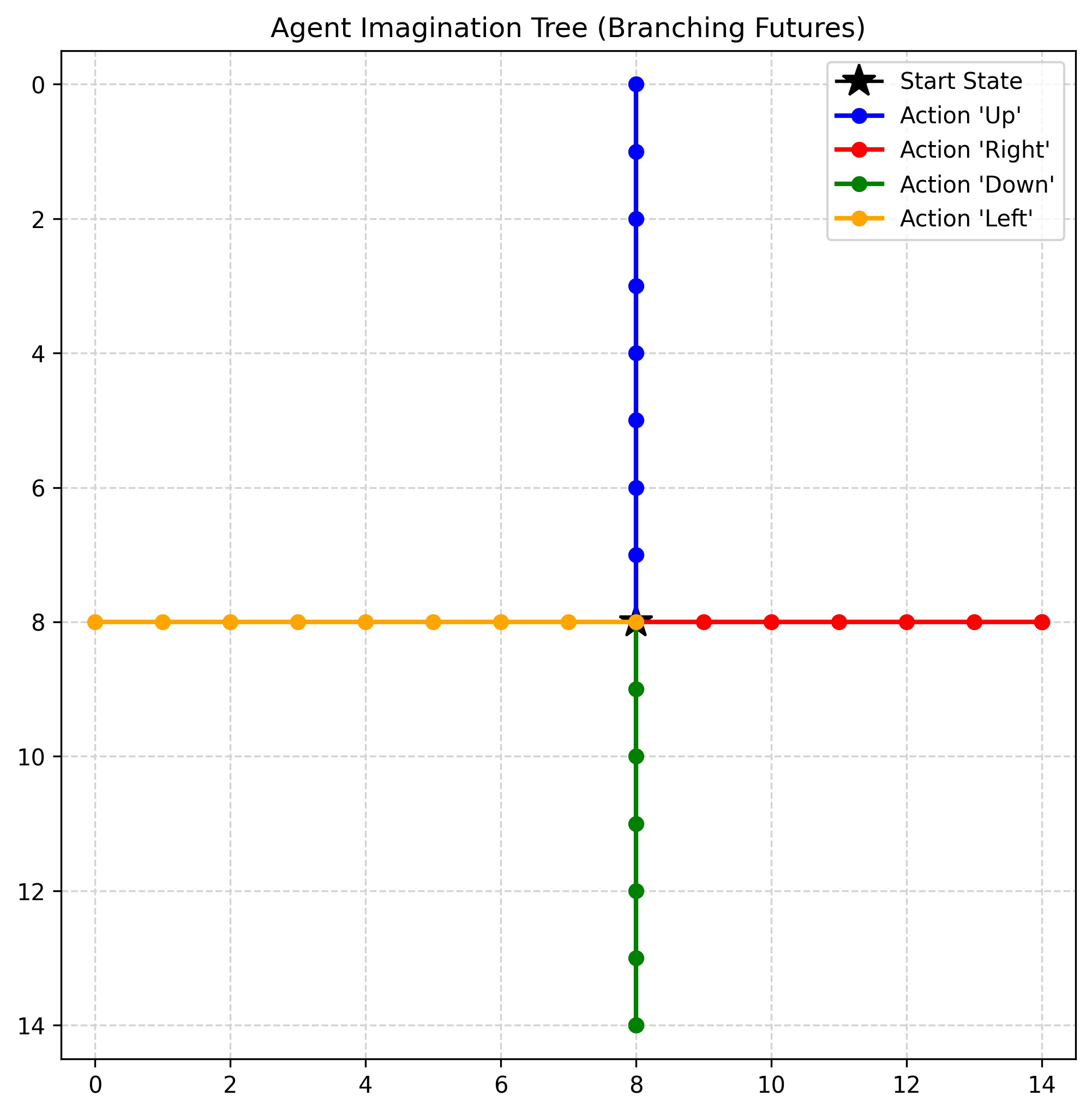}
    \caption{Agent Imagination Tree (Branching Futures). Starting from an initial state (black star), the Action-Conditioned Spatial Predictor unrolls deterministic, orthogonal branches across the latent grid for all four available actions. The network organically maps the exact empirical visitation distribution without hallucinating invalid transitions or dead nodes.}
    \label{fig:imagination_tree}
\end{figure}

\subsection{Factorized Ontologies and Dimensionality Reduction}

\begin{figure}[htbp]
    \centering
    \includegraphics[width=\textwidth]{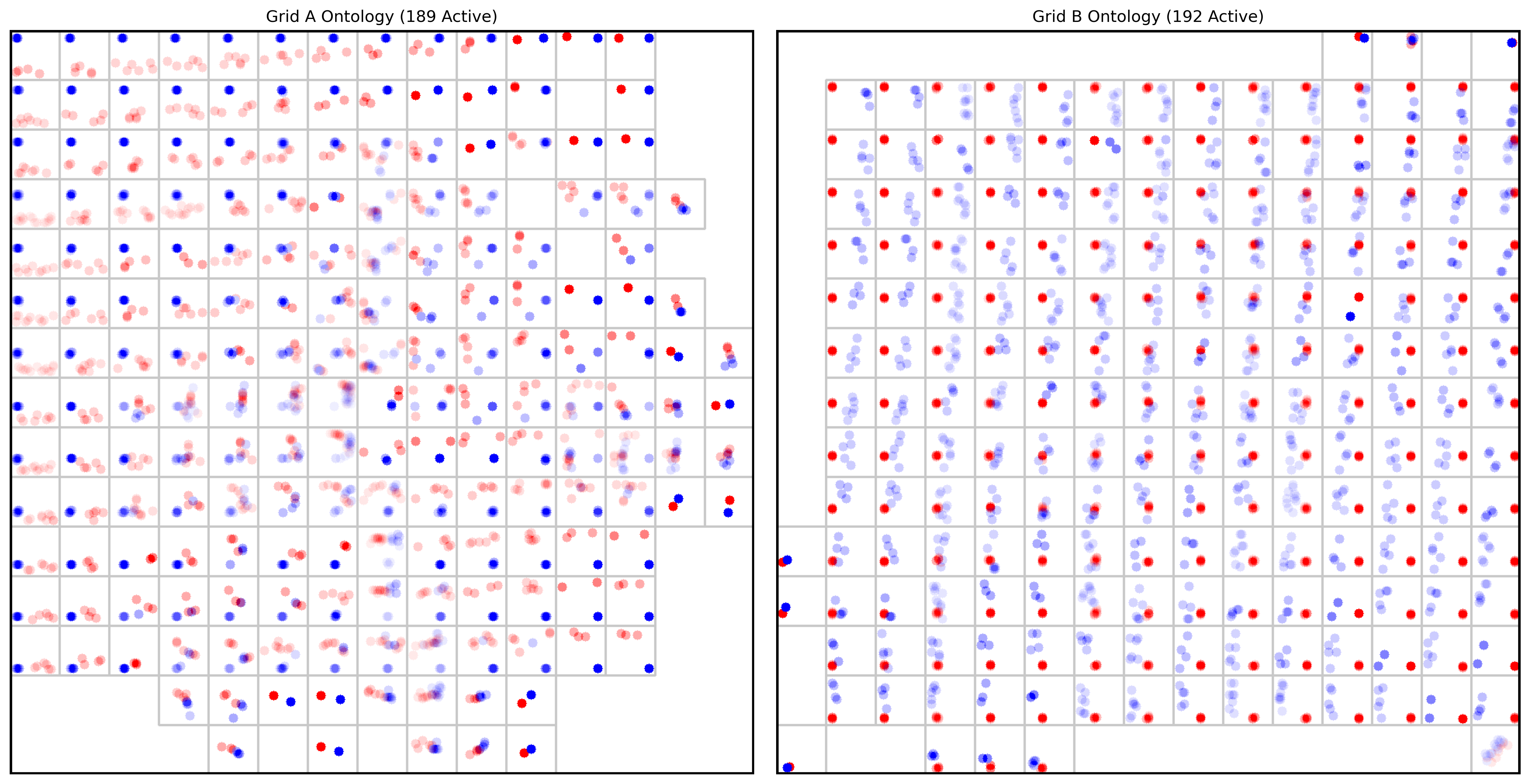}
    \caption{Dual-channel separation of a multi-entity environment. The GNWM autonomously allocates independent semantic maps to individual objects.}
    \label{fig:factorization}
\end{figure}

In the compositional paradigm, the dual-channel GNWM successfully factorized the visual scene. Grid A actively tracked the red ball while Grid B exclusively tracked the blue ball, as seen in Figure 6. By computing thermodynamics independently, the architecture successfully bypasses the combinatorial state-space explosion.

\subsection{Cognitive Topologies and Abstract Semantics}

\begin{figure}[htbp]
    \centering
    \includegraphics[width=0.5\textwidth]{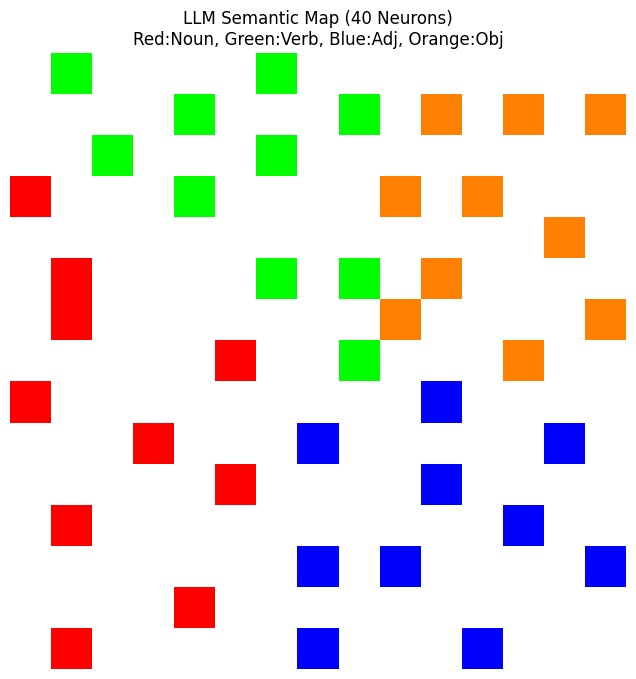}
    \caption{Grounded topological organization of abstract sequence data. The network autonomously clusters non-spatial concepts by causal grammatical category.}
    \label{fig:semantics}
\end{figure}

In the abstract sequence task, Figure 7 shows that the GNWM utilized exactly 40 active neurons to represent the 40 unique words, clustering them rigidly by grammatical category (Noun, Verb, Adjective, Object). This proves the architecture organically discovers abstract causal rules and constructs a grounded, geometric ontology of concepts beyond spatial kinematics.

\section{Conclusion}
We introduced the Global Neural World Model (GNWM), demonstrating that it is possible to achieve robust discrete topological quantization without relying on heuristic quantization loss or asymmetric gradient routing. By isolating geometric entropy constraints to a retinotopic encoder and utilizing a spatial transition predictor, the GNWM forms an interpretable and self-stabilizing state space. 

Most critically, our ablation studies across physical kinematics and abstract grammatical sequences define the exact operational boundaries of the architecture. The GNWM is not strictly a spatial physics engine; it is a causal discovery model. The thermodynamic transition loss organically builds discrete geometric maps of continuous data sequences governed by predictable rules, failing only when confronted with pure stochastic noise or unmapped vanishing entities. The ability to arrest manifold drift, factorize dual-object scenes, and generate deterministic decision trees makes the GNWM a highly capable foundation for action-conditioned planning.

We successfully deployed the fixed Gaussian convolution ($\sigma$) across a $30 \times 30$ grid ($D = 900$). Establishing whether these spatial ``neighborhood'' concepts break down at massive scale (e.g., $D = 4096$ for ViT-Huge), or if they require adaptive smoothing techniques, remains a necessary empirical pursuit before scaling the topological space to handle high-resolution, real-world visual environments.

\section*{Acknowledgements}
The author acknowledges the use of Google's Gemini as an interactive computational research assistant during the preparation of this manuscript. The AI was utilized to assist with code optimization for the visualization of experimental results, structural copyediting of the draft, and LaTeX typesetting. The author conceptualized the framework, conducted the experiments, and assumes full accountability for the scientific validity and conclusions of the work.

\appendix
\section{Combinatorial Generalization and 1D Circular Topologies}
\label{app:tsp}

While the primary text of this paper evaluates the Global Neural World Model (GNWM) within the context of spatial kinematics and temporal Joint-Embedding Predictive Architectures (JEPAs), the underlying thermodynamic equilibrium is fundamentally a generalized causal discovery engine. To demonstrate that the architecture's capacity for discrete topological quantization extends beyond spatial physics, we evaluate the GNWM on a classic combinatorial routing benchmark: the Traveling Salesman Problem (TSP).

\subsection{Historical Context: Elastic Nets and Self-Organizing Maps}
The approach of utilizing continuous topological deformations to solve discrete combinatorial problems has a rich history. The foundational \textit{Elastic Net} algorithm [9, Durbin \& Willshaw, 1987] conceptualized the TSP as a physical ``rubber band'' gradually expanding to incorporate a set of 2D coordinates. Concurrently, early applications of Kohonen's Self-Organizing Maps (SOMs) were adapted to solve the TSP by structuring the neural lattice as a 1D closed ring [11, Angéniol et al., 1988; 10, Fort, 1988]. 

However, these historical methods suffered from severe heuristic fragility. Classic SOM solvers relied on non-differentiable, hard Winner-Take-All (WTA) spikes and required meticulous tuning of dynamic learning rates to prevent ``dead neurons'' (nodes that never map to a city) or node collapse (multiple nodes mapping to the same city). Consequently, they were highly susceptible to deep local minima and tangled paths, preventing their integration into modern, end-to-end differentiable deep learning pipelines.

\subsection{The 1D Circular GNWM}
The GNWM revitalizes and solves this historical paradigm by replacing brittle heuristics with its pure, 3-term thermodynamic equilibrium. For a TSP graph of $C$ cities, we configure the GNWM's discrete topology not as a 2D Cartesian grid, but as a 1D circular array (a closed ring) of $N$ nodes, where $N > C$. 

Continuous 2D city coordinates $(x, y)$ are passed through a standard Multi-Layer Perceptron (MLP) mapping onto the $N$-dimensional space. Instead of a 2D spatial filter, Topological Smearing is applied via a 1D circular Gaussian convolution, wrapping around the boundaries of the array to form a continuous elastic loop.

The combinatorial optimization is organically solved by the tension between the GNWM's standard thermodynamic forces:
\begin{enumerate}
    \item \textbf{Expansion ($\mathcal{L}_{collapse}$):} Evaluated against a uniform constant vector $c$, this force mathematically guarantees that the probability mass expands to utilize the entire ring. This definitively solves the historical ``dead neuron'' problem, ensuring every city claims a unique locus on the topology.
    \item \textbf{Contraction ($\mathcal{L}_{WTA}$):} This forces the soft mapping to converge into hard, discrete assignments, snapping the continuous rubber band strictly to the exact city coordinates.
    \item \textbf{Topological Similarity ($\mathcal{L}_{sim}$):} Acting as the ``elastic pull,'' this penalty ensures that nodes which are adjacent on the 1D neural ring are not physically distant in the 2D Cartesian plane, organically untangling the path.

\end{enumerate}

\subsection{Dynamic Tension Decay and Emergent Routing}
To prevent the continuous topology from settling into sub-optimal ``crossed'' minima, we introduce a \textit{Dynamic Tension Decay} schedule. As the simulated annealing cools the variance ($\sigma$) of the 1D topological smearing, the scalar weight applied to the $\mathcal{L}_{sim}$ penalty is proportionally reduced. 

Initially, a high topological penalty forces the network to untangle the global macroscopic shape of the tour. As the temperature cools, reducing the elastic penalty allows the expansion and contraction forces to overpower the topological rigidity, enabling the network to violently snap to the exact coordinates of outlier cities without breaking the established sequence. 

Empirical tests on randomly generated 30-city graphs demonstrate that the GNWM consistently discovers highly optimal, continuous tours (see Figure 8). This confirms that the architecture is not merely a spatial tracking mechanism, but a fully differentiable, generalized engine capable of unrolling discrete combinatorial logic using only continuous thermodynamic gradients.
\makeatletter
\setlength{\@fptop}{0pt}
\makeatother
\begin{figure}[hbt!]
    \centering
    \includegraphics[width=\textwidth]{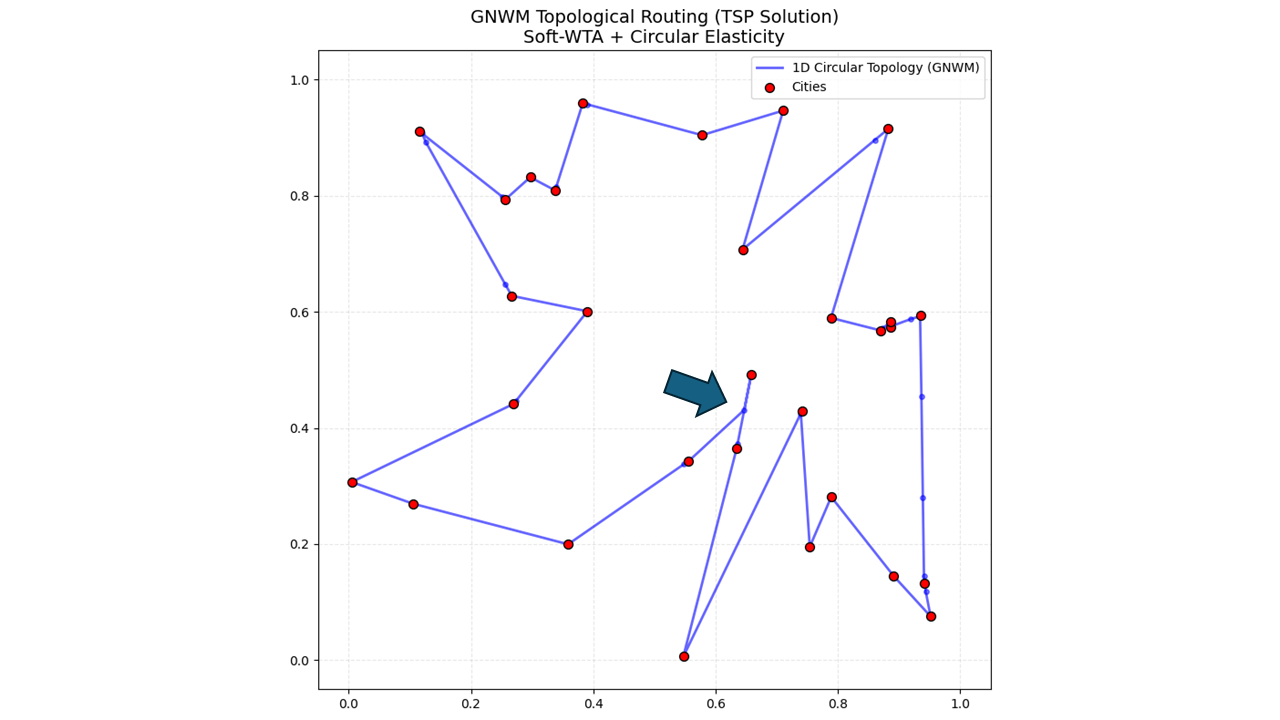}
    \caption{The GNWM For topological routing (TSP problem). Good solutions can be found by restarting algorithm with different random initial conditions but we picked this particular solution to show that Elastic Maps, even with end-to-end differentiable deep learning pipeline, are approximate algorithms as shown by the arrow. \textbf{GitHub} repository containing simulation scripts can be found \href{https://github.com/norikermiche-123/GNWM}{here.}}
    \label{fig:tsp_routing}
\end{figure}

\end{document}